\documentclass[11pt,a4paper]{article}
\usepackage[hyperref]{acl2018}
\usepackage{times}
\usepackage{latexsym}

\usepackage{url}

\aclfinalcopy

\title{AllenNLP: A Deep Semantic Natural Language Processing Platform}
\author{Matt Gardner, Joel Grus, Mark Neumann, Oyvind Tafjord, Pradeep Dasigi,\\ {\bf Nelson F. Liu, Matthew Peters, Michael Schmitz,  Luke Zettlemoyer}\\
Allen Institute for Artificial Intelligence\\
}

\date{}

\begin{document}
\maketitle
\begin{abstract}
  Modern natural language processing (NLP) research requires writing code. Ideally this code would provide a precise definition of the approach, easy repeatability of results, and a basis for extending the research. However, many research codebases bury high-level parameters under implementation details, are challenging to run and debug, and are difficult enough to extend that they are more likely to be rewritten. This paper describes AllenNLP, a library for applying deep learning methods to NLP research, which addresses these issues with easy-to-use command-line tools, declarative configuration-driven experiments, and modular NLP abstractions.  AllenNLP has already increased the rate of research experimentation and the sharing of NLP components at the Allen Institute for Artificial Intelligence, and we are working to have the same impact across the field.
\end{abstract}

\section{Introduction}

Neural network models are now the state-of-the-art for a wide range of tasks such as text classification~\cite{Howard2018FinetunedLM}, machine translation~\cite{Vaswani2017AttentionIA}, semantic role labeling~\cite{zhou2015end,he2017deep}, coreference resolution~\cite{lee2017end}, and semantic parsing~\cite{Krishnamurthy2017NeuralSP}.  However it can be surprisingly difficult to tune new models or replicate existing results. State-of-the-art deep learning models often take over a week to train on modern GPUs and are sensitive to initialization and hyperparameter settings. Furthermore, reference implementations often re-implement NLP components from scratch and make it difficult to reproduce results, creating a barrier to entry for research on many problems.

AllenNLP, a platform for research on deep learning methods in natural language processing, is designed to address these problems and to significantly lower barriers to high quality NLP research by

\begin{itemize}
    \item implementing useful NLP abstractions that make it easy to write higher-level model code for a broad range of NLP tasks, swap out components, and re-use implementations,
    \item handling common NLP deep learning problems, such as masking and padding, and keeping these low-level details separate from the high-level model and experiment definitions,
    \item defining experiments using declarative configuration files, which provide a high-level summary of a model and its training, and make it easy to change the deep learning architecture and tune hyper-parameters, and
    \item sharing models through live demos, making complex NLP accessible and debug-able.
\end{itemize}

The \href{http://allennlp.org}{AllenNLP website}\footnote{http://allennlp.org/} provides tutorials, API documentation, pretrained models, and \href{https://github.com/allenai/allennlp}{source code}\footnote{http://github.com/allenai/allennlp}. The AllenNLP platform has a permissive Apache 2.0 license and is easy to download and install via pip, a Docker image, or cloning the GitHub repository. It includes reference implementations for recent state-of-the-art models (see Section~\ref{sec:models}) that can be easily run (to make predictions on arbitrary new inputs) and retrained with different parameters or on new data.  These pretrained models have \href{http://demo.allennlp.org}{interactive online demos}\footnote{http://demo.allennlp.org/} with visualizations to help interpret model decisions and make predictions accessible to others.  The reference implementations provide examples of the framework functionality (Section~\ref{sec:features}) and also serve as baselines for future research.

AllenNLP is an ongoing open-source effort maintained by several full-time engineers and researchers at the Allen Institute for Artificial Intelligence, as well as interns from top PhD programs and contributors from the broader NLP community. It is used widespread internally for research on common sense, logical reasoning, and state-of-the-art NLP components such as: constituency parsers, semantic parsing, and word representations.  AllenNLP is gaining traction externally and we want to invest to make it the standard for advancing NLP research using PyTorch.

\section{Library Design}
\label{sec:features}

AllenNLP is a platform designed specifically for deep learning and NLP research.  AllenNLP is built on PyTorch~\cite{paszke2017automatic}, which provides many attractive features for NLP research. PyTorch supports dynamic networks, has a clean ``Pythonic'' syntax, and is easy to use.

The AllenNLP library provides (1) a flexible data API that handles intelligent batching and padding, (2) high-level abstractions for common operations in working with text, and (3) a modular and extensible experiment framework that makes doing good science easy.

AllenNLP maintains \href{https://codecov.io/gh/allenai/allennlp}{a high test coverage of over 90\%}\footnote{https://codecov.io/gh/allenai/allennlp} to ensure its components and models are working as intended.  Library features are built with testability in mind so new components can maintain a similar test coverage.

\subsection{Text Data Processing}

AllenNLP's data processing API is built around the notion of \texttt{Field}s.  Each \texttt{Field} represents a single input array to a model. \texttt{Field}s are grouped together in \texttt{Instance}s that represent the examples for training or prediction.

The \texttt{Field} API is flexible and easy to extend, allowing for a unified data API for tasks as diverse as tagging, semantic role labeling, question answering, and textual entailment.  To represent the SQuAD dataset~\cite{rajpurkar2016squad}, for example, which has a question and a passage as inputs and a span from the passage as output, each training \texttt{Instance} comprises a \texttt{TextField} for the question, a \texttt{TextField} for the passage, and a \texttt{SpanField} representing the start and end positions of the answer in the passage.

The user need only read data into a set of \texttt{Instance} objects with the desired fields, and the library can automatically sort them into batches with similar sequence lengths, pad all sequences in each batch to the same length, and randomly shuffle the batches for input to a model.

\subsection{NLP-Focused Abstractions}
\label{sec:abstractions}

AllenNLP provides a high-level API for building models, with abstractions designed specifically for NLP research. By design, the code for a model actually specifies a \emph{class} of related models. The researcher can then experiment with various architectures within this class by simply changing a configuration file, without having to change any code.

The library has many abstractions that encapsulate common decision points in NLP models.  Key examples are: (1) how text is represented as vectors, (2) how vector sequences are modified to produce new vector sequences, (3) how vector sequences are merged into a single vector.

\textbf{\texttt{TokenEmbedder:}} This abstraction takes input arrays generated by e.g. a \texttt{TextField} and returns a sequence of vector embeddings. Through the use of polymorphism and AllenNLP's experiment framework (see Section~\ref{sec:experiment-framework}), researchers can easily switch between a wide variety of possible word representations. Simply by changing a configuration file, an experimenter can choose between pre-trained word embeddings, word embeddings concatenated with a character-level CNN encoding, or even pre-trained model token-in-context embeddings~\cite{peters2017semi}, which allows for easy controlled experimentation.

\textbf{\texttt{Seq2SeqEncoder:}} A common operation in deep NLP models is to take a sequence of word vectors and pass them through a recurrent network to encode contextual information, producing a new sequence of vectors as output.  There is a large number of ways to do this, including LSTMs~\cite{hochreiter1997long}, GRUs~\cite{cho2014learning}, intra-sentence attention~\cite{cheng2016long}, recurrent additive networks~\cite{lee2017recurrent}, and many more. AllenNLP's \texttt{Seq2SeqEncoder} abstracts away the decision of which particular encoder to use, allowing the user to build an encoder-agnostic model and specify the encoder via configuration. In this way, a researcher can easily explore new recurrent architectures; for example, they can replace the LSTMs in \emph{any model} that uses this abstraction with any other encoder, measuring the impact across a wide range of models and tasks.

\textbf{\texttt{Seq2VecEncoder:}} Another common operation in NLP models is to merge a sequence of vectors into a single vector, using either a recurrent network with some kind of averaging or pooling, or using a convolutional network.  This operation is encapsulated in AllenNLP by a \texttt{Seq2VecEncoder}. This abstraction again allows the model code to only describe a \emph{class} of similar models, with particular instantiations of that model class being determined by a configuration file.

\textbf{\texttt{SpanExtractor:}} A recent trend in NLP is to build models that operate on \emph{spans} of text, instead of on \emph{tokens}.  State-of-the-art models for coreference resolution~\cite{lee2017end}, constituency parsing~\cite{Stern2017AMS}, and semantic role labeling~\cite{he2017deep} all operate in this way.  Support for building this kind of model is built into AllenNLP, including a \texttt{SpanExtractor} abstraction that determines how span vectors get computed from sequences of token vectors.

\subsection{Experimental Framework}
\label{sec:experiment-framework}

The primary design goal of AllenNLP is to make it easy to do good science with controlled experiments. Because of the abstractions described in Section~\ref{sec:abstractions}, large parts of the model architecture and training-related hyper-parameters can be configured outside of model code.  This makes it easy to clearly specify the important decisions that define a new model in configuration, and frees the researcher from needing to code all of the implementation details from scratch.

This architecture design is accomplished in AllenNLP using a HOCON\footnote{We use it as JSON with comments.  See https://github.com/lightbend/config/blob/master/HOCON.md for the full spec.} configuration file that specifies, e.g., which text representations and encoders to use in an experiment.  The mapping from strings in the configuration file to instantiated objects in code is done through the use of a \emph{registry}, which allows users of the library to add new implementations of any of the provided abstractions, or even to create their own new abstractions.

While some entries in the configuration file are optional, many are required and if unspecified AllenNLP will raise a ConfigurationError when reading the configuration.  Additionally, when a configuration file is loaded, AllenNLP logs the configuration values, providing a record of both specified and default parameters for your model.

\section{Reference Models}
\label{sec:models}

AllenNLP includes reference implementations of widely used language understanding models. These models demonstrate how to use the framework functionality presented in Section~\ref{sec:features}. They also have verified performance levels that closely match the original results, and can serve as comparison baselines for future research.

AllenNLP includes reference implementations for several tasks, including:

\begin{itemize}
    \item {\bf Semantic Role Labeling} (SRL) models recover the latent predicate argument structure of a sentence~\cite{palmer2005proposition}. SRL builds representations that answer basic questions about sentence meaning; for example, ``who'' did ``what'' to ``whom.'' The AllenNLP SRL model is a re-implementation of a deep BiLSTM model~\cite{he2017deep}. The implemented model closely matches the published model which was state of the art in 2017, achieving a F1 of 78.9\% on English Ontonotes 5.0 dataset using the CoNLL 2011/12 shared task format.

    \item {\bf Machine Comprehension} (MC) systems take an evidence text and a question as input, and predict a span within the evidence that answers the question. AllenNLP includes a reference implementation of the BiDAF MC model~\cite{seo2017bidirectional} which was state of the art for the SQuAD benchmark~\cite{rajpurkar2016squad} in early 2017.
    
    \item {\bf Textual Entailment} (TE) models take a pair of sentences and predict whether the facts in the first necessarily imply the facts in the second. The AllenNLP TE model is a re-implementation of the decomposable attention model~\cite{parikh2016decomposable}, a widely used TE baseline that was state-of-the-art on the SNLI dataset~\cite{snli:emnlp2015} in late 2016. The AllenNLP TE model achieves an accuracy of 86.4\% on the SNLI 1.0 test dataset, a 2\% improvement on most publicly available implementations and a similar score as the original paper. Rather than pre-trained Glove vectors, this model uses ELMo embeddings~\cite{Peters2018DeepCW}, which are completely character based and account for the 2\% improvement.
    
    \item A {\bf Constituency Parser} breaks a text into sub-phrases, or constituents.  Non-terminals in the tree are types of phrases and the terminals are the words in the sentence. The AllenNLP constituency parser is an implementation of a minimal neural model for constituency parsing based on an independent scoring of labels and spans ~\cite{Stern2017AMS}.  This model uses ELMo embeddings~\cite{Peters2018DeepCW}, which are completely character based and improves single model performance from 92.6 F1 to 94.11 F1 on the Penn Tree bank, a 20\% relative error reduction.
\end{itemize}

AllenNLP also includes a token embedder that uses pre-trained ELMo~\cite{Peters2018DeepCW} representations.  ELMo is a deep contextualized word representation that models both complex characteristics of word use (e.g., syntax and semantics) and how these uses vary across linguistic contexts (in order to model polysemy).  ELMo embeddings significantly improve the state of the art across a broad range of challenging NLP problems, including question answering, textual entailment, and sentiment analysis.

Additional models are currently under development and are regularly released, including semantic parsing~\cite{Krishnamurthy2017NeuralSP} and multi-paragraph reading comprehension~\cite{Clark2017SimpleAE}. We  expect the number of tasks and reference implementations to grow steadily over time.  The most up-to-date list of reference models is maintained at http://allennlp.org/models.

\section{Related Work}

Many existing NLP pipelines, such as Stanford CoreNLP~\cite{manning2014stanford} and spaCy\footnote{https://spacy.io/}, focus on predicting linguistic structures rather than modeling NLP architectures.  While AllenNLP supports making predictions using pre-trained models, its core focus is on enabling novel research. This emphasis on configuring parameters, training, and evaluating is similar to Weka~\cite{Witten1999DataMP} or Scikit-learn~\cite{scikit-learn}, but AllenNLP focuses on cutting-edge research in deep learning and is designed around declarative configuration of model architectures in addition to model parameters.

Most existing deep-learning toolkits are designed for general machine learning~\cite{bergstra2010theano,yu2014introduction,chen2015mxnet,abadi2016tensorflow,neubig2017dynet}, and can require significant effort to develop research infrastructure for particular model classes. Some, such as Keras~\cite{chollet2015keras}, do aim to make it easy to build deep learning models.  Similar to how AllenNLP is an abstraction layer on top of PyTorch, Keras provides high-level abstractions on top of static graph frameworks such as TensorFlow.  While Keras' abstractions and functionality are useful for general machine learning, they are somewhat lacking for NLP, where input data types can be very complex and dynamic graph frameworks are more often necessary.

Finally, AllenNLP is related to toolkits for deep learning research in dialog~\cite{miller2017parlai} and machine translation~\cite{klein2017opennmt}. Those toolkits support learning general functions that map strings (e.g. foreign language text or user utterances) to strings (e.g. English text or system responses). AllenNLP, in contrast, is a more general library for building models for any kind of NLP task, including text classification, constituency parsing, textual entailment, question answering, and more.

\section{Conclusion}

The design of AllenNLP allows researchers to focus on the high-level summary of their models rather than the details, and to do careful, reproducible research.  Internally at the Allen Institute for Artificial Intelligence the library is widely adopted and has improved the quality of our research code, spread knowledge about deep learning, and made it easier to share discoveries between teams. AllenNLP is gaining traction externally and is growing an open-source community of contributors~\footnote{See GitHub stars and issues on https://github.com/allenai/allennlp and mentions from publications at https://www.semanticscholar.org/search?q=allennlp.}. The AllenNLP team is committed to continuing work on this library in order to enable better research practices throughout the NLP community and to build a community of researchers who maintain a collection of the best models in natural language processing.

\bibliography{acl2018}
\bibliographystyle{acl_natbib}

\end{document}